\pdfoutput=1

\documentclass[11pt]{article}

\usepackage{acl}

\usepackage{times}
\usepackage{latexsym}

\usepackage[T1]{fontenc}

\usepackage[utf8]{inputenc}

\usepackage{microtype}

\usepackage[utf8]{inputenc}
\usepackage{booktabs}
\usepackage{graphicx}
\usepackage{CJK}
\usepackage{multirow}
\usepackage{color}
\usepackage{verbatim}
\usepackage{url}
\usepackage{enumitem}
\usepackage{amsmath}
\usepackage{diagbox}
\usepackage{flushend,cuted}
\usepackage{bbm}
\usepackage{xcolor}
\usepackage{xspace}

\newcommand{\kaiti}[1]{\begin{CJK*}{UTF8}{gkai} #1 \end{CJK*}}

\newcommand{\dataset}{FGraDA\xspace}

\title{\dataset: A Dataset and Benchmark for Fine-Grained Domain Adaptation in Machine Translation}

\author{
    Wenhao Zhu\textsuperscript{\normalfont 1},
    Shujian Huang\textsuperscript{\normalfont 1\thanks{$^*$ Corresponding author.}},
    Tong Pu\textsuperscript{\normalfont 1}, 
    Pingxuan Huang\textsuperscript{\normalfont 2}\\
    {\bf Xu Zhang\textsuperscript{\normalfont 3}},
    {\bf Jian Yu\textsuperscript{\normalfont 3}}, 
    {\bf Wei Chen\textsuperscript{\normalfont 3}},
    {\bf Yanfeng Wang\textsuperscript{\normalfont 3}} \and
    {\bf Jiajun Chen\textsuperscript{\normalfont 1}} \\
    \textsuperscript{1}National Key Laboratory for Novel Software Technology, Nanjing University, China \\
    \textsuperscript{2}University of Michigan, USA \space
    \textsuperscript{3}Sogou Inc. Beijing, China \\
    \texttt{\small{\{zhuwh, putong\}@smail.nju.edu.cn, \{huangsj, chenjj\}@nju.edu.cn}, pxuanh@umich.edu} \\
    \texttt{\small{\{zhangxu216526, yujian216093, chenweibj8871, wangyanfeng\}@sogou-inc.com}}
}

\begin{document}
\maketitle

\begin{abstract}
Previous research for adapting a general neural machine translation (NMT) model into a specific domain usually neglects the diversity in translation within the same domain,
which is a core problem for domain adaptation in real-world scenarios.
One representative of such challenging scenarios is to deploy 
a translation system for a conference with a specific topic, e.g., global warming or coronavirus, where there are usually extremely less resources due to the limited schedule.
To motivate wider investigation in such a scenario, we present a real-world fine-grained domain adaptation task in machine translation (\dataset). 
The \dataset dataset consists of Chinese-English translation task for four sub-domains of information technology: autonomous vehicles, AI education, real-time networks, and smart phone. 
Each sub-domain is equipped with a development set and test set for evaluation purposes.
To be closer to reality, \dataset does not employ any in-domain bilingual training data but provides bilingual dictionaries and wiki knowledge base, which can be easier obtained within a short time.  
We benchmark the fine-grained domain adaptation task and present in-depth analyses showing that there are still challenging problems to further improve the performance with heterogeneous resources. 
\end{abstract}

\section{Introduction}
Recent years have witnessed the great thrive in neural machine translation~\citep{sutskever2014sequence, bahdanau2015align, vaswani2017attention}. 
These neural-network-based models are wildly successful when there is abundant parallel data for training.
However, in most real-world scenarios, the amount of data in a specific domain is limited. 
Therefore, domain adaptation becomes a popular topic that aims at adapting translation models in the general domain (or a source domain) to a target domain~\citep{luong2015stanford, freitag2016fast,chu2017empirical, barone2017regularization, michel2018extreme, vilar2018learning, hu2019domain, zhao2020knowledge}. 

We notice that current research of domain adaptation usually considers very broad target domains.
E.g., the popular dataset OPUS\footnote{\url{http://opus.nlpl.eu}}~\citep{tiedemann2012parallel} is tested for the following domains: law, medical, information technology, Koran, and subtitles~\citep{koehn2017six}.
There are still strong diversities within each domain. 
For example, the subtitles domain contains subtitles from action movies, political movies, Sci-Fi movies, etc. 

\begin{table}[!tbp]
    \footnotesize
    \centering
    \scalebox{0.9}{
    \begin{tabular}{l|c}
    \toprule
    Domain &  translations around the word \kaiti{``卡''} \\
    \midrule
    Autonomous Vehicles & ...  the wheel is {\color{red} \textit{stuck}} and you can't ... \\
    \midrule
    AI Education & ... some of these math {\color{red} \textit{card}} games ...\\
    \midrule
    Real-Time Networks & ... how to fix video {\color{red} \textit{stuttering}} ...\\
    \midrule
    Smart Phone & ... find your {\color{red} \textit{SIM card}} slot and ...\\
    \bottomrule
    \end{tabular}
    }
    \caption{An example where the Chinese word \kaiti{``卡''} have different translations (shown in red italics fonts) in different sub-domains of information technology.}
    \label{tab:domain_keywords}
\end{table}

\begin{table*}[!tbp]
\centering
\footnotesize
\scalebox{1}{
\begin{tabular}{c|p{2.3cm}<{\centering}|p{3cm}<{\centering}|p{2.3cm}<{\centering}|p{2.3cm}<{\centering}}
\toprule
     Domain              &  Dictionary         & Wiki knowledge base  &  Development set   &   Test set     \\
     & (items) & (wiki pages) & (sent. pairs) & (sent. pairs) \\ 
\midrule
Autonomous Vehicles (AV) &  275                &  116,381             &  200 &  605                           \\
AI Education (AIE)       &  270                &  195,339             &  200 & \hspace{-0.3cm} 1,309         \\
Real-Time Networks (RTN) &  360                &  111,101             &  200 & \hspace{-0.3cm} 1,303                  \\
Smart Phone (SP)         &  284                &  \hspace{0.1cm}90,337             &  200 &  750                    \\
\bottomrule
\end{tabular}
}
\caption{Main statistics of our dataset. 
We report the number of items for the dictionary, the number of wiki pages in the extracted wiki knowledge base, and the number of sentence pairs of development set and test set, respectively.}
\label{tab:statistics}
\end{table*}

We suggest that there are fine-grained sub-domains within these coarse domains.
The sentences or words in different sub-domains may have different language phenomena, which requires a fine-grained treatment.
As shown in Table \ref{tab:domain_keywords}, at the word level, the same Chinese word \kaiti{``卡''} may correspond to different English translations in different fine-grained information technology~(IT) domains.
Capturing semantic diversity may be hard in traditional coarse domain adaptation but is often needed in real-world scenarios, such as translation services for a specific conference or translating a technical monograph.

To make the situation even worse, adaptation to these fine-grained domains often face challenges as a low-resource scenario, because there are limited time and budget, e.g. for the translation service provider, to collect data (especially parallel data) in the fine-grained domain. Specific research may be needed to explore other heterogeneous resources which are more available.
The hierarchy between coarse and fine-grained domains may raise other research challenges as well.

In this paper, we introduce a novel and challenging dataset for fine-grained domain adaptation in machine translation, namely \dataset\footnote{All the \dataset dataset resources is released at \url{https://github.com/OwenNJU/FGraDA}} (Section~\ref{sec:dataset}). 
\dataset includes Chinese-English translation tasks on four fine-grained domains: autonomous vehicles (AV), AI education (AIE), real-time networks (RTN), and smart phone (SP), which are all sub-domains of IT. The development and test sets used for evaluation are collected and anonymized from real-world conferences.

For each fine-grained domain, no parallel training data is provided, as in the real-world scenarios parallel training data is expensive to obtain. 
For the purpose of adaptation, we provide heterogeneous but more available resources: bilingual dictionaries and wiki knowledge base.

We conduct benchmark experiments to facilitate further comparison (Section~\ref{sec:benchmark}), as well as in-depth analyses to show translation errors (Section~\ref{sec:challenges}) and interesting research challenges (Section~\ref{sec:rethinking}).
Please note that this fine-grained domain adaptation problem is so challenging that we are here only presenting and benchmarking this task and calling for attention and solutions.

\section{Related Work}
Besides OPUS, there are other popular datasets for domain adaptation, including IWSLT\footnote{\url{https://wit3.fbk.eu}}~\citep{cettoloEtAl:EAMT2012} and
ASPEC\footnote{\url{http://lotus.kuee.kyoto-u.ac.jp/ASPEC/}}~\citep{nakazawa2016aspec}.
The IWSLT corpus is usually used as a single target domain of technical talks. But it comprises a collection of Ted talks coming from very diverse areas, such as biology, chemistry, psychology, etc. 
The ASPEC corpus includes scientific papers from several different disciplines, such as physics, earth science, agriculture, etc. And these domains are not as specific as those for fine-grained scenarios.
Due to the substantial diversity within each domain, these current datasets cannot be directly used to simulate the fine-grained setting. 

We notice that novel fine-grained datasets always set advance a research field in natural language processing, such as the fine-grained task in entity recognition~\citep{hovy-etal-2006-ontonotes} and sentiment analysis~\citep{pontiki-etal-2014-semeval}. 
We hope our dataset also offers a stepping stone for further research of fine-grained domain adaptation.

\section{\dataset Dataset}
\label{sec:dataset}
The task of \dataset is to improve the translation quality for each fine-grained domain with the provided heterogeneous resources, such as bilingual dictionaries and wiki knowledge bases, so that the methodology could be used to improve real-world applications in a quick and economical way.

The dataset is built from one representative of real-world fine-grained adaptation scenarios, which is to provide translation services (i.e., simultaneous interpretation) for specific international conferences~\citep{gu-etal-2017-learning}.
These conferences mainly focus on very specific topics, such as global warming, coronavirus, etc.
However, it is difficult and costly for translation service providers to obtain massive in-domain parallel data for each specific conference in a short time.
Therefore it is potentially useful to explore methods of adapting the NMT system with other more available resources in such a predicament.

We present the fine-grained domains and the resources in the following subsections (main statistics are shown in Table~\ref{tab:statistics}).

\begin{table*}[htbp]
\footnotesize
\centering
\scalebox{0.9}{
\begin{tabular}{l|l|l|l}
\toprule
         \hspace{0.6cm}Autonomous Vehicles           & \hspace{1.2cm}AI Education              & \hspace{0.3cm}Real-Time Networks     & \hspace{1cm}Smart Phone\\  
\midrule
        \kaiti{自动驾驶} -    self-driving           &  \kaiti{知识检索} - knowledge retrieval & \kaiti{直播} - live streaming        &  \kaiti{蓝牙} - bluetooth               \\
        \kaiti{超声波雷达} - ultrasonic radar        &  \kaiti{虚拟教学} - virtual teaching    & \kaiti{丢包} - packet loss           &  \kaiti{高动态范围成像} - HDR           \\
        \kaiti{车道协同} - lane coordination         &  \kaiti{脑电图} - EEG                   & \kaiti{网络地址转换} - NAT           &  \kaiti{焦外} - bokeh                   \\
        \kaiti{激光雷达} - LiDAR                     &  \kaiti{聊天机器人} - chatbot           & \kaiti{传输层} - transport layer     &  \kaiti{帧率} - fps                \\
        \kaiti{行人检测} - pedestrian detection      &  \kaiti{机器学习} - machine learning    & \kaiti{延迟} - latency               &  \kaiti{蜂窝网络} - cellular network    \\
\bottomrule
\end{tabular}
}
\caption{Examples of the annotated bilingual dictionary}
\label{tab:bidict}
\end{table*}

\subsection{Fine-Grained Domains}

We select four real-world conferences as representatives to construct the dataset. 
\begin{figure}
    \centering
    \includegraphics[width=0.45\textwidth]{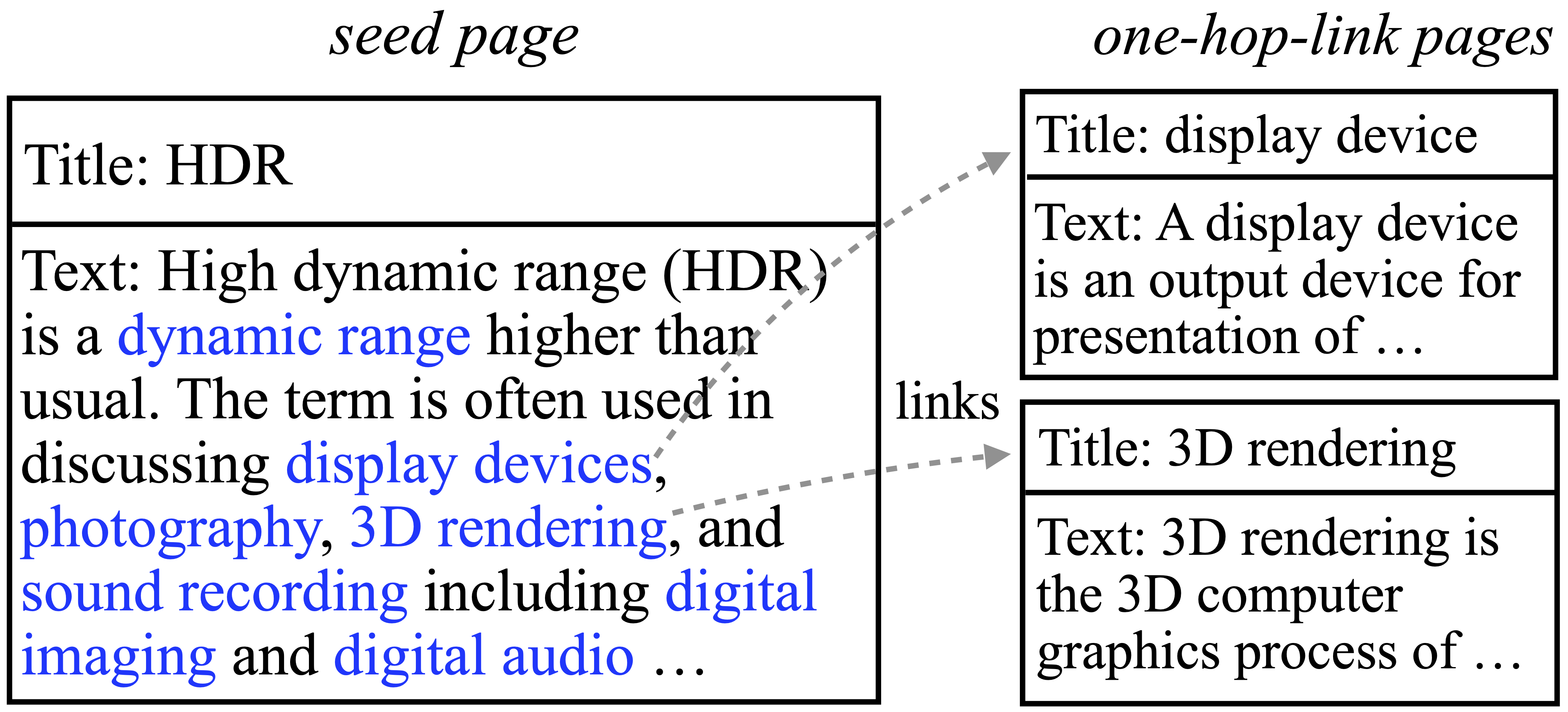}
    \caption{Illustration of the wiki knowledge base provided in our dataset. The seed page contains words (in blue font) that have links pointing to other pages.}
    \label{fig:wiki}
\end{figure}
Each conference\footnote{The four conferences are the Global AI and Robotics Conference (CCF-GAIR2019), the GIIS China Education Industry Innovation Summit (GIIS2019),  the Real-Time Internet Conference (RTC2019), and Apple Events (held in 2018 and 2019). } is organized for a particular topic of IT, namely autonomous vehicles, AI education, real-time networks, and smart phone, which could be seen as four fine-grained domains.

These fine-grained domains has obviously different topics. But they may still share some common words or language phenomenons of the IT domain, showing an interesting characteristic inside the domain hierarchy, which are less studied in previous researches.

\subsection{Bilingual Dictionary}
Compared to parallel data, a set of domain-specific keywords or phrases may be much easier or cheaper to obtain. 
Translating these keywords or phrases can provide important, domain-specific word-level correspondences between the two languages and act as the starting point of the adaptation process.
As the dictionary will later be used for retrieving wiki resources, the quality and domain relevance of the dictionary is quite important.

\begin{table}[htbp]
    \footnotesize
    \centering
    \scalebox{0.9}{
        \begin{tabular}{c|c|c}
        \toprule
        Domain              &   seed pages & one-hop-link pages  \\
        \midrule
        Autonomous Vehicle &    19,277 / 490      &   97,104 / 1,522        \\
        AI Education       &    35,615 / 636      &   \hspace{-0.17cm}159,724 / 1,536        \\
        Real-Time Networks  &    17,930 / 565      &   93,171 / 1,386        \\
        Smart Phone        &    15,944 / 452      &   74,393 / 1,736        \\
        \bottomrule
        \end{tabular}
    }
    \caption{Detailed statistics of our wiki knowledge base.
    In each cell, the left numbers correspond to the number of extracted pages and the right numbers correspond to the average number of words contained in one page.}
    \label{tab:wiki}
\end{table}

Therefore, we manually build a small set of domain-specific keywords/phrases for each domain as bilingual dictionaries. Table \ref{tab:bidict} shows some examples. 
To make sure that the selection and translation of domain-specific words are reliable, we have all the dictionary items checked by linguistic experts. 

\subsection{Wiki Knowledge Base}
Wikipedia is an useful resource for machine translation~\citep{halek2011named, wu-etal-2019-machine}.
As we have obtained a manually checked in-domain dictionary, it is convenient to retrieve wikipages with the given dictionary.
Because aligned wikipedia pages in different languages are not always available, we only use the wikipages in the target language as our resources.

More specifically, we first collect English wikipages\footnote{\url{https://dumps.wikimedia.your.org/enwiki/20200701/}} containing annotated dictionary keywords in their titles. These pages are closely related to the fine-grained domain (later mentioned as seed pages).
Since each wikipage naturally contains words and links pointing to other related wikipages (illustrated in Figure \ref{fig:wiki}), we also leverage these structural knowledge to collect more information. We collect the wikipages directly linked by links in the seed pages (later mentioned as one-hop-link pages). This one-hop constraint makes sure that these pages are relevant to our domain.

Our final knowledge base for each domain consists of both seed pages and one-hop-link pages. 
Statistics are shown in Table~\ref{tab:wiki}.
Our knowledge base not only contain rich monolingual resources, but also have additional structural knowledge, e.g. link relations, which may be of potential usage.

Compared with the existing wiki knowledge base, such as DBpedia\footnote{\url{https://wiki.dbpedia.org}}, which is in the form of instance-properties pairs, our wiki knowledge base costs less time to build, contains more information~(in the text of the pages), and is more closely related to the fine-grained target domains. 

\subsection{Development and Test Set}
To evaluate performances on the \dataset task, we also collect and label parallel data. 
We collect 70 hours of real-world audio recordings from the four conferences mentioned above, transcript the audio recordings with in-house tools, filter out domain-irrelevant sentences\footnote{Note that filtering is conducted as we only concern translation performance on the domain related part. }, and annotate them into 4,767 parallel sentence pairs (Table~\ref{tab:statistics}). 
Data desensitization is then conducted as post-editing to hide human names and company names in the annotation data to protect privacy. 
Each of the above steps is consulted with linguistic experts, so the labeling process is expensive, which is why a large amount of parallel data is no easy to obtain.

We split annotated data in each domain into two parts: 200 sentence pairs as the development set, and the rest as the test set.
We do understand the size of the development set is relatively small for a typical large scale machine translation system, but improving the translation under this condition may be a practical problem. In contrast, the test sets are larger for better and consistent evaluation.


\section{Benchmarks}
\label{sec:benchmark}

\subsection{Notations}
NMT systems typically generate a target language sentence $\mathbf{y}\textrm{=}\{y_1, y_2, \cdots, y_{|\mathbf{y}|}\}$ given a source language sentence $\mathbf{x}\textrm{=}\{x_1, x_2, \cdots, x_{|\mathbf{x}|}\}$ in an end-to-end fashion. The translation probability distribution is factorized as: 
\begin{equation}
p_{\theta}(\mathbf{y}|\mathbf{x}) = \prod \limits_{i=1}^{|\mathbf{y}|} p(y_i | \mathbf{x}, \mathbf{y}_{<i};\theta)   , 
\end{equation}
where $y_i$ is the current predicted token; $\mathbf{y}_{<i}$ is the previous predicted tokens and $\theta$ is the parameters of the NMT model. 
The model can be trained by minimizing the loss on the training set $D$:
\begin{equation}
\mathcal{L}(D;\theta) = \sum_{(\mathbf{x},\mathbf{y})\in D} -\log p_{\theta}(\mathbf{y}|\mathbf{x}) .
\end{equation}

For domain adaptation, we denote the general domain parallel data as $D_{\textrm{g}}=\{(\mathbf{x}_{\textrm{g}}, \mathbf{y}_{\textrm{g}})\}$, the in-domain parallel data as $D_{\textrm{in}}=\{(\mathbf{x}_{\textrm{in}}, \mathbf{y}_{\textrm{in}})\}$ and the in-domain monolingual data in target language as $M_{\textrm{in}}=\{\mathbf{y}_{\textrm{in}}\}$, where the subscript is used to identify the corresponding domain.

\subsection{Domain Adaptation Approaches} 
We briefly categorize and discuss existing domain adaptation approaches according to the resources they employ.
\paragraph{Using (pseudo) parallel data:}
Given in-domain parallel data, the most popular domain adaptation approach is fine-tuning~\citep{luong2015stanford}, where a general domain model trained on $D_{\textrm{g}}$ is continuously trained on $D_{\textrm{in}}$ by minimizing the loss:
\begin{equation}  
    \mathcal{L_{FT}}(D_{\textrm{in}};\theta_\textrm{in}, \theta_\textrm{g}) = \sum_{(\mathbf{x}_{\textrm{in}},\mathbf{y}_{\textrm{in}})\in D_{\textrm{in}}}-\log p_{\theta_\textrm{in}}(\mathbf{y}_{\textrm{in}}|\mathbf{x}_{\textrm{in}}),
\end{equation}
where $\theta_\textrm{g}$ and $\theta_\textrm{in}$ is the parameters of the general and adapted model, respectively. $\theta_\textrm{in}$ is initialized by $\theta_\textrm{g}$.


When parallel data is not available, leveraging monolingual data with back-translation~(BT) is an alternative.
BT translates monolingual data in the target language $M_{\textrm{in}}$ back to the source language to construct pseudo parallel data $\hat{D}_{\textrm{in}}=\{(\hat{\mathbf{x}}_{\textrm{in}}, \mathbf{y}_{\textrm{in}})\}$, and uses $\hat{D}_{\textrm{in}}$ as an augmentation for fine-tuning~\citep{sennrich2016improving, hoang2018iterative}.


Parallel data is the exact type of resource that NMT systems are trained from, thus is useful for adaptation. But high quality parallel data is expensive to obtain in our scenario.

\paragraph{Using dictionaries:} 
Grid beam search (GBS)~\citep{hokamp-liu-2017-lexically} is proposed to incorporate dictionaries into NMT at decoding time, which extends beam search to allow the inclusion of pre-specified lexical constraints.
Large-scale bilingual dictionaries can also be treated as pseudo bitext for 
fine-tuning~\citep{kothur2018document, thompson2019hablex} at training time,
or for building word-by-word translation to generate pseudo bitext~\citep{hu2019domain}. 

\dataset provide a small but high quality set of domain specific dictionary. Using it as decoding constraints might be a reasonable choice. However, compared to general lexical constraints, domain specific constraints are harder to generate because these constraints are often rare in the general domain. To balance the constraints and the generation process, we re-weight the log-likelihood during constrained generation steps:
\begin{equation}
\begin{split}
\textrm{score}(\hat{\mathbf{y}}, \mathbf{x}) =  -\prod \limits_{i=1}^{|\hat{\mathbf{y}}|} [\mathcal{I}(\hat{y}_i\notin \mathcal{C})\log p(\hat{y}_i | \mathbf{x}, \hat{\mathbf{y}}_{<i};\theta)\\
+ (1-w)\mathcal{I}(\hat{y}_i\in \mathcal{C})\log p(\hat{y}_i | \mathbf{x}, \hat{\mathbf{y}}_{<i};\theta)] ,
\end{split}
\end{equation}
where $\hat{\mathbf{y}}$ is the generated hypothesis sentence and $\mathcal{C}$ is the constraint set.
Moreover, rather than selecting hypotheses only from beams where all the constraints are satisfied, we select highest scored hypotheses from all beams, which enables the model to ignore constraints that are harmful to the generation process.


\paragraph{Using Knowledge Base:} Bilingual knowledge base could be used for extracting bilingual lexicons \cite{zhao2020knowledge}. To our best knowledge, there is not previous attempts in exploring domain related information in monolingual wikipages, as provided in \dataset.
As one of the first attempts, we simply take all sentences from seed pages as $M_{\textrm{in}}$ and apply back-translation.

\subsection{Benchmark Systems}


We implement the following systems as benchmark baselines.

\textbf{$\textrm{Base}$:} Directly using a Transformer~\citep{vaswani2017attention} trained on $D_{\textrm{g}}$ on the target domains without any adaptation.

\textbf{$\textrm{Dict}_\textrm{GBS}$:}
Performing constrained decoding~\cite{hokamp-liu-2017-lexically} for $\textrm{Base}$ with in-domain dictionary. We implement a weighted version described in the previous sub-section. The weight is selected on the development set (see Section \ref{ssec:m_dict} for discussions about the weight).

\textbf{$\textrm{Dict}_\textrm{FT}$:} 
Fine-tuning the $\textrm{Base}$ model on the in-domain dictionary~\citep{kothur2018document}.

\textbf{$\textrm{Wiki}_\textrm{BT}$:}
Using sentences of wiki seed pages for back-translation with the $\textrm{Base}$ model ~\citep{sennrich2016improving}.

\textbf{$\textrm{Wiki}_\textrm{BT}\textrm{+}\textrm{Dict}_\textrm{GBS}$:}
Applying constrained decoding on $\textrm{Wiki}_\textrm{BT}$ model.

\subsection{Experiment Settings}

\paragraph{General Domain:} We use WMT-CWMT-17 Chinese-English dataset (9 million sentence pairs) as the general domain data and train $\textrm{Base}$ model with newsdev2017\footnote{We report validation performance in Appendix A.} as the development set. 
For back translation, the backward model is also trained on WMT-CWMT-17 zh-en dataset.

\paragraph{Data Processing:} We use the open-source toolkit sentence\_splitter\footnote{\url{https://github.com/mediacloud/sentence-splitter}} to split paragraphs in wikipages into sentences.
We use the script in \textit{moses}\footnote{\url{https://github.com/moses-smt/mosesdecoder}} and \textit{jieba}\footnote{\url{https://github.com/fxsjy/jieba}} to tokenize the English and Chinese corpus, respectively. 
Byte-pair encoding~\citep{sennrich2016neural} is applied with 32k merge operations. 

\paragraph{Implementation Details:}
All the models are implemented with an open source tool NJUNMT\footnote{\url{https://github.com/whr94621/NJUNMT-pytorch}} and follow the architecture of transformer-base~\citep{vaswani2017attention}.
Adam is used as the optimizer and Noam as the learning rate scheduler.
We set 8k warm-up steps and a maximum learning rate as 9e-4.
We train the $\textrm{Base}$ model on 4 Tesla V100, which takes three days. 
The batch size is 3000 tokens, and the update circle is 10.
Beam size is set as 5.

In all fine-tuning experiments, we choose learning rate from \{1e-6, 5e-7, 1e-7, 5e-8\} according to model's BLEU score on the development set.

We report detokenized case-insensitive BLEU scores calculated with \textit{mteval-v13.pl}\footnote{\url{https://github.com/moses-smt/mosesdecoder/blob/master/scripts/generic/mteval-v13a.pl}}.

\subsection{Benchmark Results}
\begin{table}[t]
    \footnotesize
    \centering
    \begin{tabular}{l|cccc|c}
    \toprule
         Model                            & AV                    & AIE          & RTN                & SP & Avg.  \\
    \midrule
    \textbf{$\textrm{Base}$}              & 34.0                  & 31.1         & 16.6               & 22.9        & 26.2    \\
    \textbf{$\textrm{Dict}_\textrm{GBS}$} & 34.5                  & 31.1         & 17.0               & 23.0        & 26.4\\
    \textbf{$\textrm{Dict}_\textrm{FT}$}  & 34.0                  & 31.1         & 16.7               & 22.9        & 26.2\\                  
    \textbf{$\textrm{Wiki}_\textrm{BT}$}  & 34.8                  & 31.8         & 16.8               & 23.4        & 26.7       \\
    \textbf{$\textrm{Wiki}_\textrm{BT}\textrm{+}\textrm{Dict}_\textrm{GBS}$}  & \textbf{35.1}         & \textbf{31.9}& \textbf{17.2}      & \textbf{23.6} & \textbf{27.0}\\
    \bottomrule
    \end{tabular}
    \caption{Translation results (BLEU scores) on four fine-grained domains. Bold text identifies the best result among the benckmark result. ``Avg'' denotes the average results across four domains.}
    \label{tab:bleu_scores}
\end{table}

\begin{table*}[t]
\newcommand{\tabincell}[2]{\begin{tabular}{@{}#1@{}}#2\end{tabular}}
    \centering
    \footnotesize
    \scalebox{1}
    {
    \begin{tabular}{l|p{13.5cm}}
    \toprule
    \multicolumn{2}{l}{Type \uppercase\expandafter{\romannumeral1}: mistranslating domain-specific words} \\
    \midrule
    Source & \kaiti{如果 你 想 直接 从 一个 浏览器 发送信息 到 另 一个 浏览器 ， 唯一 的 办法 就是 使用 {\color{blue} 网页 即时 通信 技术} 。}  \\
    \midrule
    Hypothesis & \tabincell{l}{If you want to send messages directly from one browser to another, the only way to do so is  to use \\ {\color{red} \textit{web instant communication technology}}.} \\
    \midrule
    Reference & The only way in which you can send a message directly from one browser to the other is using {\color{red} \textit{WebRTC}}. \\
    \midrule
    \midrule
    \multicolumn{2}{l}{Type \uppercase\expandafter{\romannumeral2}: misunderstanding common words with domain specific meaning} \\
    \midrule
    Source & \kaiti{左边 是 相对 {\color{blue} 卡} 很多 ， 右边 是 相对 流畅 ， 也 有 {\color{blue}卡顿} ， 但是 总体 上 流畅 度 有 巨大 的 提升。}  \\
    \midrule
    Hypothesis & \tabincell{l}{On the left is a lot of relative {\color{red} \textit{cards}}, on the right is relatively fluid, also there is {\color{red} \textit{Carton}}, but overall fluency \\ has a great increase.} \\
    \midrule
    Reference & \tabincell{l}{The left is relatively {\color{red} \textit{stutter}}. The right is relatively smooth, and there are {\color{red} \textit{stutters}}, but the overall \\ fluency is greatly improved.}\\
    \midrule
    \midrule
    \multicolumn{2}{l}{Type \uppercase\expandafter{\romannumeral3}: under-translating the source sentence } \\
    \midrule
    Source & \kaiti{ 但是我们也注意到，这种{\color{blue}送达模式}在以前非常重要。} \\
    \midrule
    Hypothesis & \tabincell{l}{But we also note that this {\color{red} \textit{service pattern}} was important in the past.} \\
    \midrule
    Reference & \tabincell{l}{However, we also notice that although this {\color{red} \textit{delivery mode}} used to be very important.} \\
    \bottomrule
    \end{tabular}
    }
    \caption{Typical types of errors and examples. Source fragments that are wrongly translated are shown in blue font. The error in the hypothesis and the correct translation in the reference are shown in red and italic font.}
    \label{tab:case}
\end{table*}

The benchmark results are presented in Table \ref{tab:bleu_scores}.
For each domain, $\textrm{Dict}_\textrm{GBS}$ and $\textrm{Wiki}_\textrm{BT}$ improve the baseline model to some extent, while
$\textrm{Dict}_\textrm{FT}$ barely brings any improvements.
With both resources, $\textrm{Wiki}_\textrm{BT}\textrm{+}\textrm{Dict}_\textrm{FT}$ achieves the best performance among all systems. These results demonstrate the effectiveness of the heterogeneous resources. We will use this best model as the adapted model for further analyses.

However, the translation quality does not improve as greatly as reported in other research~\citep{freitag2016fast, chu2017empirical};
the performance on real-time networks and smart phone are much lower than other two domains, showing the diversity and difficulty of these fine-grained domains. 
In the following sections, we conduct further analyses to better understand the difficulty and challenges in this scenario.

\section{Translation Analyses}
\label{sec:challenges}
\begin{figure}[t]
    \centering
    \includegraphics[width=0.42\textwidth]{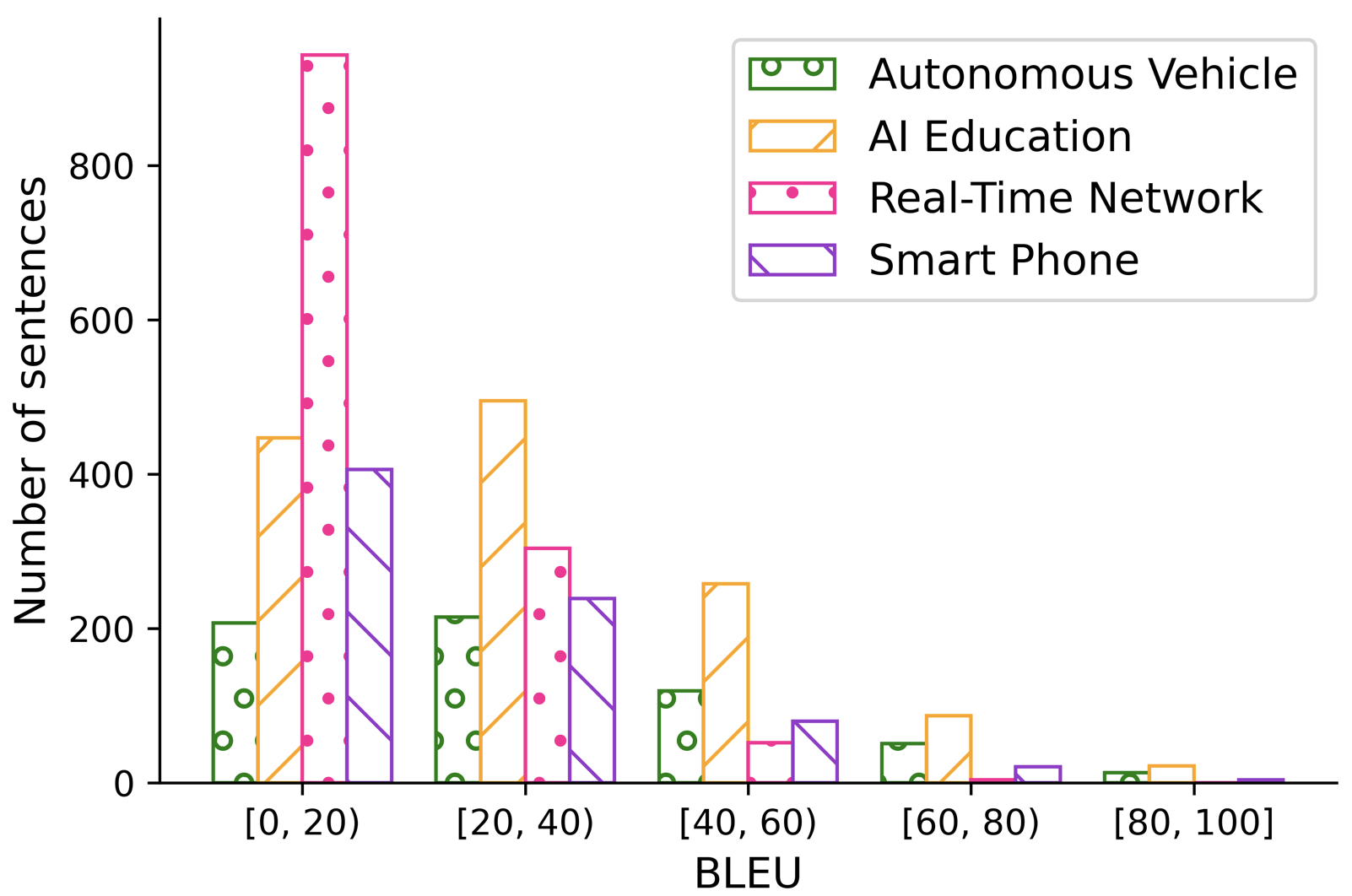}
    \caption{Distribution of sentence BLEU scores on four domains.}
    \label{fig:bleu_dist}
\end{figure}

For analysis, we translate the four test sets with the best adapted system, i.e. $\textrm{Wiki}_\textrm{BT}\textrm{+}\textrm{Dict}_\textrm{GBS}$.
We compute the BLEU score for each sentence and plot their distributions in Figure \ref{fig:bleu_dist}.
It is obvious that the translation performance on a large portion of test sentences is not satisfactory, e.g. under 20, leaving a large room for improvement.

We carefully go through generated translations of test sentences in each domain to analyze translation errors. 
We find that there are three typical types of error, which are challenging and require more attention. 
The error types and examples are presented in Table~\ref{tab:case}.

\paragraph{Mistranslating domain-specific words:} 
Some words in the specific domain rarely appear in general domain training data, which increases the difficulty for the model to translate them.
For example, the Chinese phrase \kaiti{``网页即时通信技术''} is expected to be translated as ``WebRTC'', which is a domain-specific word in real-time networks. 
Sadly, the system generates an wordy and incorrect translation (the first case in Table \ref{tab:case}).

\paragraph{Misunderstanding common words with domain specific meaning:} 
Some frequent words are endowed with a new meaning under the context of a specific domain.
Taking the Chinese word \kaiti{``卡''} as an example, it means ``card'' or ``carton'' in most cases in the general domain; but in the networking domain, it means ``stutter'' (as shown in the second case in Table \ref{tab:case}).

\paragraph{Under-translating the source sentence:} 
Part of the domain-related source sentence information is missing after translation. 
For example, in the third case in Table \ref{tab:case}, the translation model can not completely capture the semantic meaning of \kaiti{``送达模式''} and only translates it as ``service pattern'' rather than ``delivery mode''. 

All the above errors are closely related to domain specific problems, which brings interesting challenges: how to solve these problems with limited resources.

\section{Remaining challenges}
\label{sec:rethinking}
\subsection{Mining from the Dictionary}\label{ssec:m_dict}
The domain dictionary contains accurate translation knowledge about the domain specific words. We conduct analyses to see whether these knowledge are properly used. We count the occurrence of the dictionary item in the source sentence and the translation results, and compute the accuracy on each test set (Table \ref{tab:dict_coverage}).

\begin{table}[t]
    \centering
    \footnotesize
    \scalebox{1}
    {
    \begin{tabular}{l|c|c|c|c}
    \toprule
    Model & AV     & AIE    & RTN   & SP \\
    \midrule
    \textbf{$\textrm{Base}$}                                        & 63.04  & 57.81  & 65.86 & 59.42 \\
    \textbf{$\textrm{Dict}_\textrm{GBS}$}                           & \textbf{65.84}  & 59.69  & 76.94 & 61.85 \\
    \textbf{$\textrm{Wiki}_\textrm{BT}$}                            & 63.93  & 59.38  & 67.30 & 58.97  \\
    \textbf{$\textrm{Wiki}_\textrm{BT}\textrm{+}\textrm{Dict}_\textrm{GBS}$} & \textbf{65.84} & \textbf{64.22}  & \textbf{87.84} & \textbf{63.07}  \\
    \bottomrule
    \end{tabular}
    }
    \caption{The translation accuracy (\%) of items in the domain dictionary. Bold font marks the model with the highest accuracy.}
    \label{tab:dict_coverage}
\end{table}

Compared with $\textrm{Base}$ model, using these dictionary items as constraints $\textrm{Dict}_\textrm{GBS}$ improves the translation accuracy. It is interesting to see that the constrained decoding achieves even better performance when using together with the domain related monolingual data $\textrm{Wiki}_\textrm{BT}$. Considering that finetuning the dictionary does not lead to any improvement (Table~\ref{tab:bleu_scores}), it is likely that the wiki data provides sentence-level context for the domain specific words, which encourages the generation of these words. 
However, even with the adapted models, a large portion of items are still mis-translated.

\begin{figure}[t]
    \centering
    \includegraphics[width=0.40\textwidth]{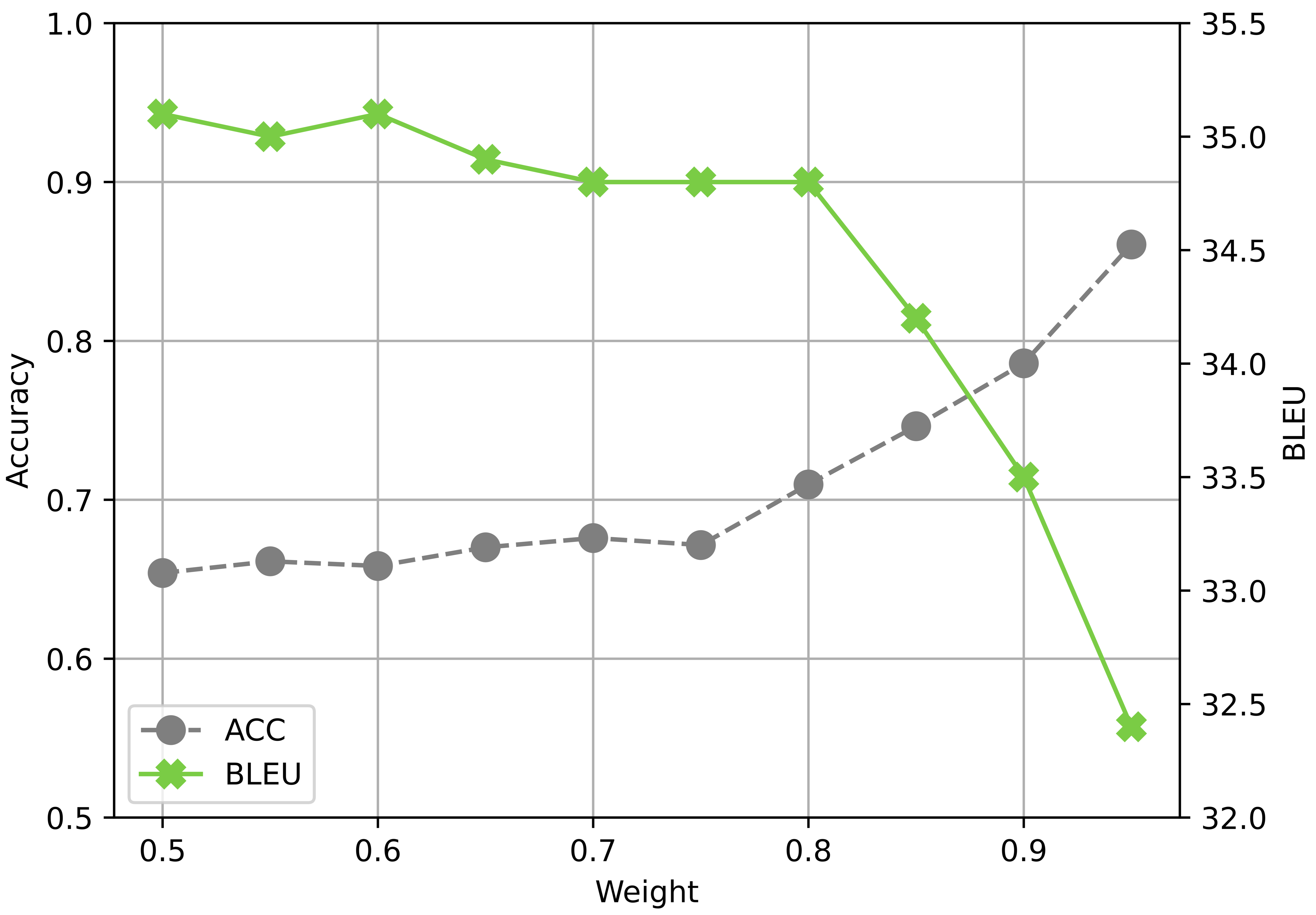}
    \caption{Dictionary words translation accuracy and BLEU scores w.r.t different weights in grid beam search on AV test set.}
    \label{fig:gbs}
\end{figure}

We set the weight of constraints at different values, and plot the translation accuracy of dictionary items and overall BLEU scores for the adapted system ($\textrm{Wiki}_\textrm{BT}\textrm{+}\textrm{Dict}_\textrm{GBS}$). 
As shown in Figure \ref{fig:gbs}, although higher weight ensures more domain specific words are translated, BLEU score drops significantly.

The results indicate that simply forcing the models to generate infrequent in-domain words is not sufficient. 
Therefore, better methods for generating these domain words are still worth exploring.

\subsection{Mining from Wiki Knowledge Base}

Another possible reason of the poor generation of domain specific words is that they are rare in the general domain, so the model is not able to learn their proper representation. We notice that resources from Wiki knowledge base may contain rich structural knowledge that may help the system to learn the representations, i.e. to ``understand'' these words.

On one hand, many wikipages contain definition for their title words, which is usually the first sentence in the page. 
For example, ``High dynamic range (HDR) is a dynamic range higher than usual'' gives the definition of ``HDR'' (illustrated in Figure \ref{fig:wiki}). It might be possible to utilize such knowledge to better understand domain-specific words in sentences.

On the other hand, most wikipages have links pointing to other words or phrases that are closely related to the current title word (also illustrated in Figure \ref{fig:wiki}).
These words, e.g. dynamic range and 3D rendering, photography, may also help to learn the representation of the title word (e.g. HDR).

Moreover, compared to the limited amount of manually labeled dictionary items, wiki knowledge base also contains much more domain related words in the one-hop-link pages. 
It is interesting to explore effective methods to utilize these knowledge in the future.

\subsection{Mining from the Domain Hierarchy}
\begin{figure}[t]
    \centering
    \includegraphics[width=0.40\textwidth]{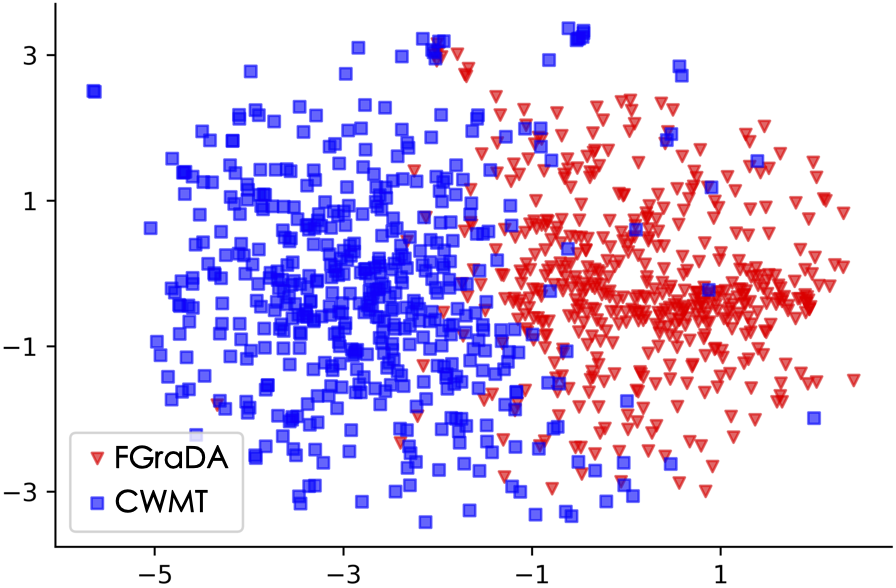}
    \caption{Visualization of sentences from \dataset and CWMT.}
    \label{fig:cwmt-fdmt}
\end{figure}

To get more insight into the relation between domains of \dataset, we use BERT~\citep{devlin2018bert} to encode source sentences in the test sets and visualize them with t-SNE~\citep{maaten2008visualizing}.
The visualizations of CWMT  (general domain) and \dataset (IT domain) data are shown in Figure \ref{fig:cwmt-fdmt}. 
The two distributions are almost separated, which means that \dataset is quite different from the general domain.

\begin{figure}[t]
    \centering
    \includegraphics[width=0.40\textwidth]{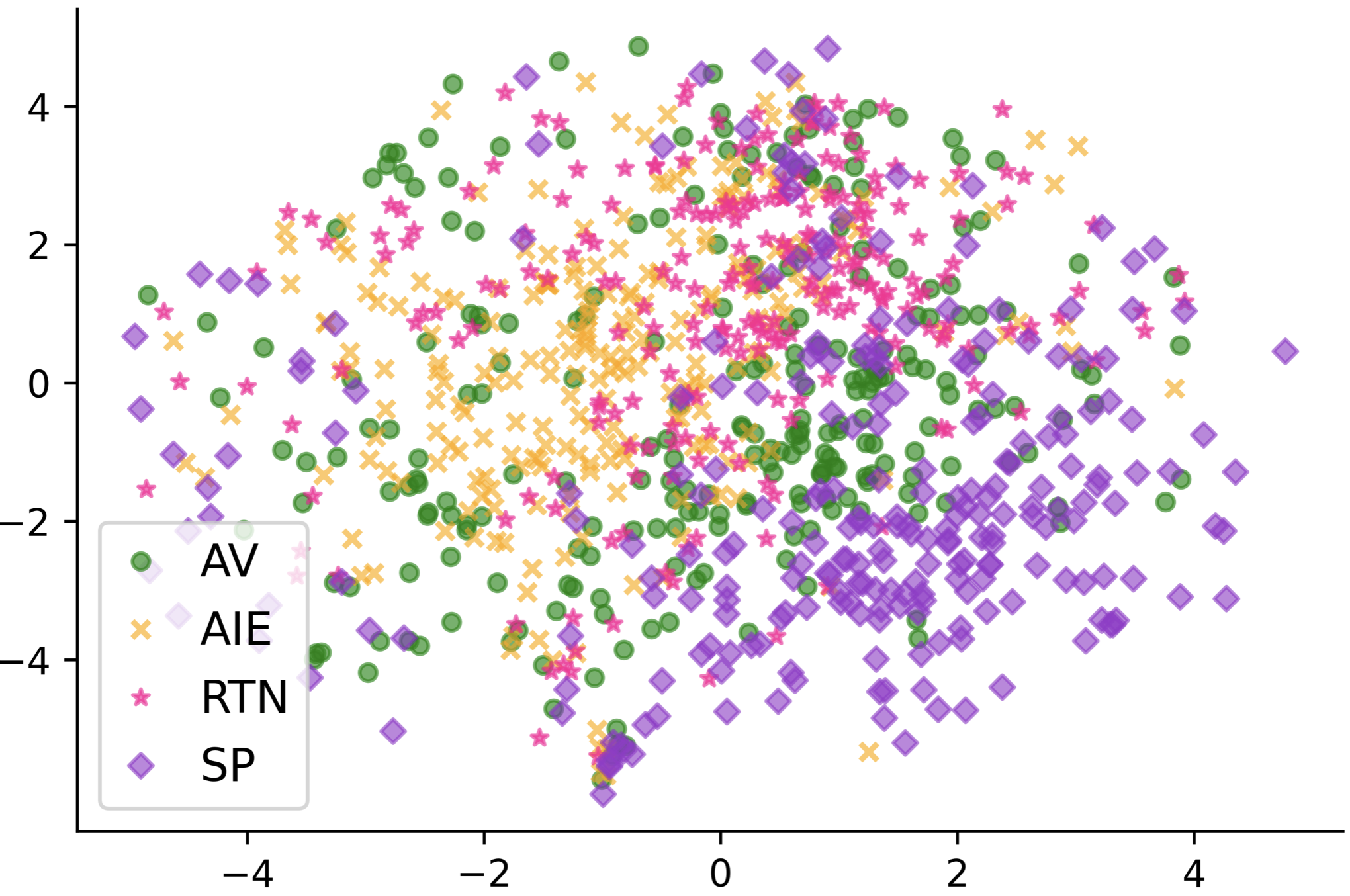}
    \caption{Visualization of sentences from the four fine-grained domains in \dataset.}
    \label{fig:fdmt}
\end{figure}

Then we plot sentences of the four fine-grained domains in Figure \ref{fig:fdmt}. 
Different from the situation in Figure \ref{fig:cwmt-fdmt}, the four IT sub-domains have more overlaps, showing they are closely related. However, each domain still presents a unique distribution.

To quantify the diversity and relation among four sub-domains, We use the adapted model for each domain to translate the test sets of other three domains (Table \ref{tab:diag}). 

For each domain, the performance varies according to the adapted data, which demonstrates the diversity among different fine-grained domains. The best performance almost distributes on the diagonal, showing that exact resource for the fine-grained domain are most helpful in most cases.
Also, the performance on some domains are different. We guess that is because some fine-grained domain, e.g. AIE and AV, may also include sentences related to other fine-grained domains. 

\begin{table}[t]
    \footnotesize
    \centering
    \begin{tabular}{l|cccc}
        \toprule
        \diagbox[width=1.8cm]{Adapt}{Test}     
                    &   AV      &  AIE       &  RTN         &  SP          \\
        \midrule
        AV          &  35.1     &  31.0      & 16.7         & 23.1    \\
        
        AIE         &  35.0     &  \textbf{31.9}      & 16.9         & 23.3   \\
        
        RTN         &  \textbf{35.2}     &  \textbf{31.9}      & \textbf{17.2}         & 23.4   \\
    
        SP          &  \textbf{35.2}     &  \textbf{31.9}      & 16.9         & \textbf{23.6}         \\
        \bottomrule
    \end{tabular}
    \caption{The performance on all four test sets. 
    Each line represents a model adapted to a specific domain.}
    \label{tab:diag}
\end{table}

To further understand the relation between different fine-grained domain, we also analyze the overlap ratio of other provided resources, i.e. dictionary and wiki knowledge base. Quantitative results in Table \ref{tab:sim} show small but consistent overlap ratios of dictionary and wiki knowledge base resources (around 5\% and 20\%, respectively), which also indicates a close relation between these domains.

\begin{table}[t]
    \footnotesize
    \centering
    \begin{tabular}{l|cccc}
        \toprule
        Resource           &   AV        &  AIE   &  RTN            &  SP          \\
        \midrule
        Dict-AV          &    -       &  5.45           &  5.82           & 3.27 \\
        Dict-AIE         & 5.56       &       -         &  3.70           & 1.45  \\
        Dict-RTN         & 4.44       &  2.78           &     -           & 5.83 \\
        Dict-SP          & 3.17       &  1.41           &  7.39           &    -     \\
        \midrule 
        Wiki-AV         &  -         & 24.29           &  23.72          & 16.03  \\
        Wiki-AIE         &  15.29     & -               &  15.20          & \hspace{+0.2cm}9.98 \\
        Wiki-RTN         &  24.85     & 26.73           &   -             & 17.44 \\
        Wiki-SP          &  20.65     & 21.59           &  21.45          & - \\
        \bottomrule
    \end{tabular}
    \caption{Overlap ratio (\%) of dictionary and wikipage resources between different domains.
    }
    \label{tab:sim}
\end{table}

From the above results, it might be beneficial to leverage resources from other related sub-domains for adaptation.
The hierarchy of coarse and fine-grained domains may also raise other interesting research topics.

\section{Conclusion}
This paper introduces the first fine-grained domain-adaptation dataset for machine translation, \dataset, which presents a real-world problem.
We benchmark the dataset and show the needs for fine-grained adaptation. 
We find that the NMT model usually mis-translates domain-specific words, misunderstands common words with domain-specific meaning, and under-translates the source sentence in our task.
Our analyses also show that the provided hetergeneous resources may contains useful information for the adaptation. However, current adaptation methods cannot effectively utilize these resources.
The challenging problems of \dataset encourages further exploration of dictionaries, wiki knowledge base, which might be more available than in-domain parallel data. It is also interesting to further explore the domain hierarchy as well.

\section{Acknowledgments}
We would like to thank the anonymous reviewers for their insightful comments. Shujian Huang is the corresponding author. This work is supported by National Science Foundation of China (No. U1836221, 6217020152), National Key R\&D Program of China (No. 2019QY1806) .


\bibliographystyle{acl_natbib}
\bibliography{acl}

\newpage
\appendix
\section{Appendix}
We report our $\textrm{Base}$ model's performance on newsdev2017 and newsdev2018 in Table\ref{tab:newsdev}. 
The comparison result shows that our transformer-base is comparable with other implementation~\citep{wang2018towards}.
\begin{table}[htbp]
    \footnotesize
    \centering
    \scalebox{0.9}
    {
    \begin{tabular}{l|c|c}
        \toprule
                      &   newsdev2017         &  newsdev2018            \\
        \midrule
        Ours          &   22.9                    &  24.3                   \\
        
        Wang et al.   &   -                   &  24.4              \\
        \bottomrule
    \end{tabular}
    }
    \caption{Evaluation results. All BLEU scores reported in this table is computed by \textit{multi-bleu.pl}\protect\footnotemark.}
    \label{tab:newsdev}
\end{table}




\end{document}